# LLMs for Survey Text Analysis – A Performance Comparison Between Humans and GPT-5 on Inductive Content Analysis

Leonardo Bergmann[1,2], Renata Gheorghiu[3], Ana Gvritishvili[4], Alex Mican[5], Chris Stewart[6], Topias Tolonen-Weckström[7]

[1]Vienna Doctoral School in Cognition, Behavior and Neuroscience (VDS CoBeNe), University of Vienna, Austria

[2]Department of Cognition, Emotion, and Methods in Psychology, Faculty of Psychology, University of Vienna, Austria

[3]International Institute for Advanced Studies in Psychotherapy and Applied Mental Health, Babeș-Bolyai University, Cluj-Napoca, Romania

[4]School of Education, Humanities and Social Sciences, International Black Sea University, Georgia

[5]Department of Economic and Social Policy, Faculty of Economics, Prague University of Economics and Business

[6]Department of Geography, University of Galway, Ireland

[7]Department of Mathematics, Uppsala University, Sweden

## Author Note

Correspondence concerning this article should be addressed to Leonardo Bergmann.

Email: leonardo.bergmann@univie.ac.at

## Abstract

Large language models (LLMs) are increasingly used to support text analysis in qualitative research, yet evidence on their performance in inductive content analysis remains limited. This study compares human and LLM-based inductive coding of open-ended survey responses from 903 answers across six variables from a European PhD student survey. Five human coders performed inductive content analysis following a standardized coding scheme, while an LLM (GPT-5.4) conducted the same task using an established prompting procedure. Agreement between human and LLM outputs was assessed using the Adjusted Rand Index (ARI). Results showed an alignment between humans and the LLM, with ARI values of 0.61 for coding and 0.54 for theme generation. These values were close to the internal consistency of coding and theme results within humans (ARI = 0.68) and the LLM (ARI = 0.76). Agreement varied widely across variables, with low within-entity consistency consistently linked to low between-entity agreement, underscoring the role of data characteristics and individual performance in reliability. Overall, the findings suggest that LLMs can approximate human coding in this case-specific setting, particularly at the coding level, and may serve as a scalable support tool for inductive qualitative analysis.

## Introduction

Large language models (LLMs) have shown a variety of benefits to support research across disciplines (see Bergmann et al., 2026, for a review). Especially for textual analysis, a key qualitative research approach, the scientific community would benefit from optimizing this time consuming task with LLM-based tools (Hitch, 2023). One major textual analysis method is content analysis, which provides researchers with a systematic approach to extract content-related categories from textual data, such as chat messages or interviews, either deductively or inductively (Elo & Kyngäs, 2008). In the deductive approach, researchers categorize text data into predefined categories. In contrast, the inductive approach involves generating initial codes, typically in the form of words capturing the basic semantic content of text fragments, inductively out of the textual data. These codes are then grouped into themes (also emerging from the process and not predefined), forming representative constructs of the text that serve as the basis for further analysis. By condensing content into themes (e.g., accessibility), this method enables both qualitative and quantitative analysis of the original material, as well as facilitating computer-supported analysis (Harwood & Garry, 2003).

Many studies have already provided evidence for the reliable performance of LLMs for simple deductive coding, especially with binary categories (Bergmann et al., 2025, Chew et al., 2023; Pham et al., 2023). However, data on the inductive LLM approach and its comparison with humans is still rare. To provide further insights to the reliability of inductive content analysis the aim of this study was to compare the coding results from humans to the ones from an LLM. Specifically, we applied the LLM-supported inductive content analysis outlined by Bergmann et al. (2025) to open text survey data and calculated the cluster correlation between human and LLM.

## Methods

As a basis for a policy document regarding third-cycle education the European Students' Union (ESU) collected survey data from 2800 PhD students across Europe (ESU, 2026). Six of the survey items had an open text answer format and were evaluated with the LLM-supported inductive content analysis outlined by Bergmann et al. (2025). To investigate the reliability of this approach, 10 % of the survey responses were randomly drawn and served as a data basis for the current study.

Five human coders performed an inductive content analysis on the sample data based on the coding instruction and coding mask outlined in Appendix A. Each human coder received a coding mask containing data of one variable with open text answer format, except one human coder performing the inductive content analysis on the last two survey variables.

For the LLM-supported inductive content analysis, we accessed GPT-5.4 via the OpenAI API and let the sample data be evaluated using the procedure from Bergmann et al. (2025). The corresponding prompts and a short description of our approach are outlined in Appendix B.

To estimate the alignment between human and LLM results, we used the Adjusted Rand Index (ARI) comparing two partitions by looking at all pairs of points and checking whether each pair is placed in the same or different clusters in both partitions. Similar to correlations this index has 0 value in the case of random partition, and it is bounded above by 1 in the case of perfect agreement between two partitions. The comparisons were between the coding and theme results of humans and the LLM as well as within humans and the LLM for both outputs. Because responses could receive multiple codes, code assignments were transformed into co-assignment matrices before calculating ARI. This procedure allowed comparison between human and LLM outputs despite the multi-label nature of the coding task. The data was analysed using the *mclust* package in the statistical data analysis software *R* (Scrucca et al., 2023). Additionally to this quantitative approach we also presented the human coders with the results from the LLM and asked them to describe their impression and if they would trust the LLM to perform the task.

## Results

In total 903 survey responses resulting from the following six variables were analysed: (1) *Accessibility Increase* ($n$ = 266), (2) *Financial Situation* ($n$ = 199), (3) *Well Being* ($n$ = 173), (4) *Mobility Barriers* ($n$ = 146), (5) *Personal Experience* ($n$ = 64), and (6) *Anything Else* ($n$ = 55). The full original survey question of each variable can be found in Appendix C.

The average ARI between humans and the LLM was 0.61 for the coding results and 0.54 for the resulting themes. These values indicate a level of agreement closer to perfect concordance than to random partitioning and may be interpreted as moderate according to common interpretations of correlation values (Schober et al., 2018). The ARI values within human coders for coding and themes were 0.68, while the LLM showed comparable internal consistency of 0.76. The ARI values reported as within-entity consistency compare the alignment between coding-level and theme-level classifications generated by the same coding entity, rather than inter-coder reliability across multiple humans coding the same material. Taken together, these findings suggest that the alignment within coding entities deviates only slightly from the alignment between human coders and the LLM. This strengthens the interpretation that the observed agreement between humans and the LLM is relatively robust.

For a more detailed investigation, Table 1 presents the individual ARI values for each variable across the previously mentioned categories. A noticeable variation in ARI values can be observed, ranging from low levels of agreement (0.31) to high levels (0.89). Comparing ARI values across variables reveals that some variables consistently exhibit higher levels of alignment than others. For instance, all ARI values for the variable (2) *Financial Situation* are at or below 0.71, whereas all values for the variable (6) *Anything Else* are at or above 0.70. Additionally, the internal consistency of ARI values within humans and the LLM seems to be related with the performance of coding and themes between the coding entities. This could explain the low ARI theme value for (2) *Financial Situation* and (3) *Well Being* due to the low internal consistency of LLM for these variables. Similarly, the low internal consistency of the human for (6) *Personal Experience* could have influenced the moderate ARI values between the coding entities.

However, in general the performance according to the ARI values seemed promising which was also confirmed by the statements of the human coders. All of them indicated that the LLM coding reflects their own categorizations and that they would support using this approach to analyse further data. The full opinions on the LLM coding of each variable can be found in Appendix D. All coding results from humans and the LLM, as well as the analysis code in *R*, are openly available online (https://osf.io/rgkeh/overview?view_only=96ba2b33f3f44ec2a3fe2570257424d1).

**Table 1**

*Adjusted Rand Index Within and Between Humans and GPT-5*

| Variable | Human | LLM | Coding | Theme |
|---|---|---|---|---|
| (1) Accessibility Increase | NA | 0.86 | 0.44 | NA |
| (2) Financial Situation | 0.71 | 0.53 | 0.35 | 0.31 |
| (3) Well Being | 0.67 | 0.67 | 0.72 | 0.41 |
| (4) Mobility Barriers | 0.88 | 0.82 | 0.79 | 0.65 |
| (5) Personal Experience | 0.46 | 0.81 | 0.51 | 0.64 |
| (6) Anything Else | 0.70 | 0.89 | 0.85 | 0.70 |

*Note.* The column *Human* reports the Adjusted Rand Index (ARI) comparing coding results and themes among human coders, while the column *LLM* presents the same comparison for LLM outputs. The column *Coding* shows the ARI between human coders and the LLM for coding results, and the column *Theme* reports the corresponding comparison for themes. The NA value is attributable to missing theme coding from the individual responsible for variable (1) *Accessibility Increase*.

## Discussion

This study examined the reliability of LLM-supported inductive content analysis by comparing its outputs with human coders. Overall, the results show moderate agreement between humans and the LLM, particularly for coding (ARI = 0.61), with slightly lower alignment for themes (ARI = 0.54). Notably, this level of agreement is close to the consistency observed within human coders themselves for the alignment of coding and themes (ARI = 0.68), suggesting that differences between humans and the LLM are comparable to typical human consistency.

The lower agreement for themes indicates that while LLMs perform well in identifying and labeling basic semantic units, they are less aligned with humans when generating higher-level abstractions. This might be due to more degrees of freedom introduced by an additional categorization step. Related work shows that agreement on themes is also lower than on codings between humans and this reduction of similarity is stable across comparison groups (Bergmann et al., 2025).

A key finding is the strong variation in agreement across variables and due to the internal consistency of the coding entity. While some topics tended toward high alignment and others toward much lower alignment, low consistency within coding entities was always associated with low agreement between coding entities, highlighting the importance of data characteristics and individual performance in determining reliability.

Overall, our case-specific study suggests that LLMs are a promising tool for supporting early stages of inductive analysis, supporting similar results found by Bergmann et al. (2025). These results should not be interpreted as suggesting that LLMs replace human qualitative judgment. Rather, they indicate that LLMs may support the early, labor-intensive stages of inductive coding, while human researchers remain essential for contextual interpretation, validation, and theory-building. To overcome limitations of this study including the reliance on a single model, the absence of a clear ground truth, and codings of single humans per variable, future research should explore different models, prompting strategies, and evaluation approaches with more human coders per variable.

**Author Contributions (CRediT framework)**

Conceptualization: Leonardo Bergmann

Data curation: Leonardo Bergmann

Formal analysis: Leonardo Bergmann

Investigation: Leonardo Bergmann, Renata Gheorghiu, Ana Gvritishvili, Alex Mican, Chris Stewart, Topias Tolonen-Weckström

Methodology: Leonardo Bergmann

Project administration: Leonardo Bergmann

Resources: Leonardo Bergmann

Writing - original draft: Leonardo Bergmann

Writing - review & editing: Renata Gheorghiu, Ana Gvritishvili, Alex Mican, Chris Stewart, Topias Tolonen-Weckström

## Appendix A - Coding Instruction and Coding Mask

### Inductive content analysis Instruction

You will be given all survey responses in an Excel document (see example below). After familiarizing yourself with the survey responses by skimming through them, you will independently generate codes in the form of concise words or phrases that capture the basic semantic content of each individual response. Priority should be given to using already generated codes for categorization before creating new ones. Depending on the survey response, more than one code may be assigned where required. After coding all survey responses, you will group the resulting codes into overarching themes that represent semantically related codes. Please choose a suitable abstraction level as these themes are then quantitatively evaluated and interpreted in the position paper (not too specific, but also not too broad).

**Example for adopting inductive content analysis**

A survey response could look like “Doctoral education would be more accessible if tuition fees were reduced and more scholarships were available, especially for students from low-income backgrounds.” and might be coded as “reduced tuition fees; Increased scholarships; financial support for low-income students” (if some of these codes already exist from earlier responses you would reuse them rather than creating new ones). These codes may then be further categorized into the broader theme of “Financial Accessibility”, depending on how all the other codes look like.

**Example for the working process in Excel**

In one column you find the survey responses. You go through them line by line and generate codes in the column next to it. As different codes can apply to one survey response, you can provide multiple codes in the code column. Please separate each code with a semicolon (;), so that we can automate the process in a later state. The Excel mask for doing that working step is provided below.

**Figure A.1**

| | A | B | C |
|---|---|---|---|
| 1 | **ID** | **12. How would you increase accessibility of doctoral educa** | **Codes (separated by ; )** |
| 2 | 1 | I don't think it's a matter of accessibility, but a matter of promo | |
| 3 | 2 | I'd say having more information about the overall workload and | |
| 4 | 3 | More information and better access to funding for diverse traje | |
| 5 | 4 | Increase accessibility to doctoral education in Portugal by expa | |
| 6 | 5 | - | |
| 7 | 6 | Funding, more project based places | |

After having all survey responses coded, please try to find overarching themes and assign all codes to the corresponding themes. The Excel mask for doing that working step is provided below.

**Figure A.2**

| | A | B | C | D | E | F |
|---|---|---|---|---|---|---|
| 1 | Theme | Code1 | Code2 | Code3 | … | |
| 2 | | | | | | |
| 3 | | | | | | |
| 4 | | | | | | |
| 5 | | | | | | |

## Appendix B - LLM-Supported Inductive Content Analysis

For the inductive content analysis we will follow a two-times two-step approach. First we define codes from the text messages, and then we categorize the survey responses to the codes. Then we define categories from the codes, and then we group the codes to the categories. The following prompts represent the described steps and are used to investigate the six open-format variables from the survey. We accessed GPT-5.4 from OpenAI via their API (https://platform.openai.com/) and inserted the prompts with the survey responses into the user field. A screenshot of the platform with the used model settings can be found below.

**Figure B.1**

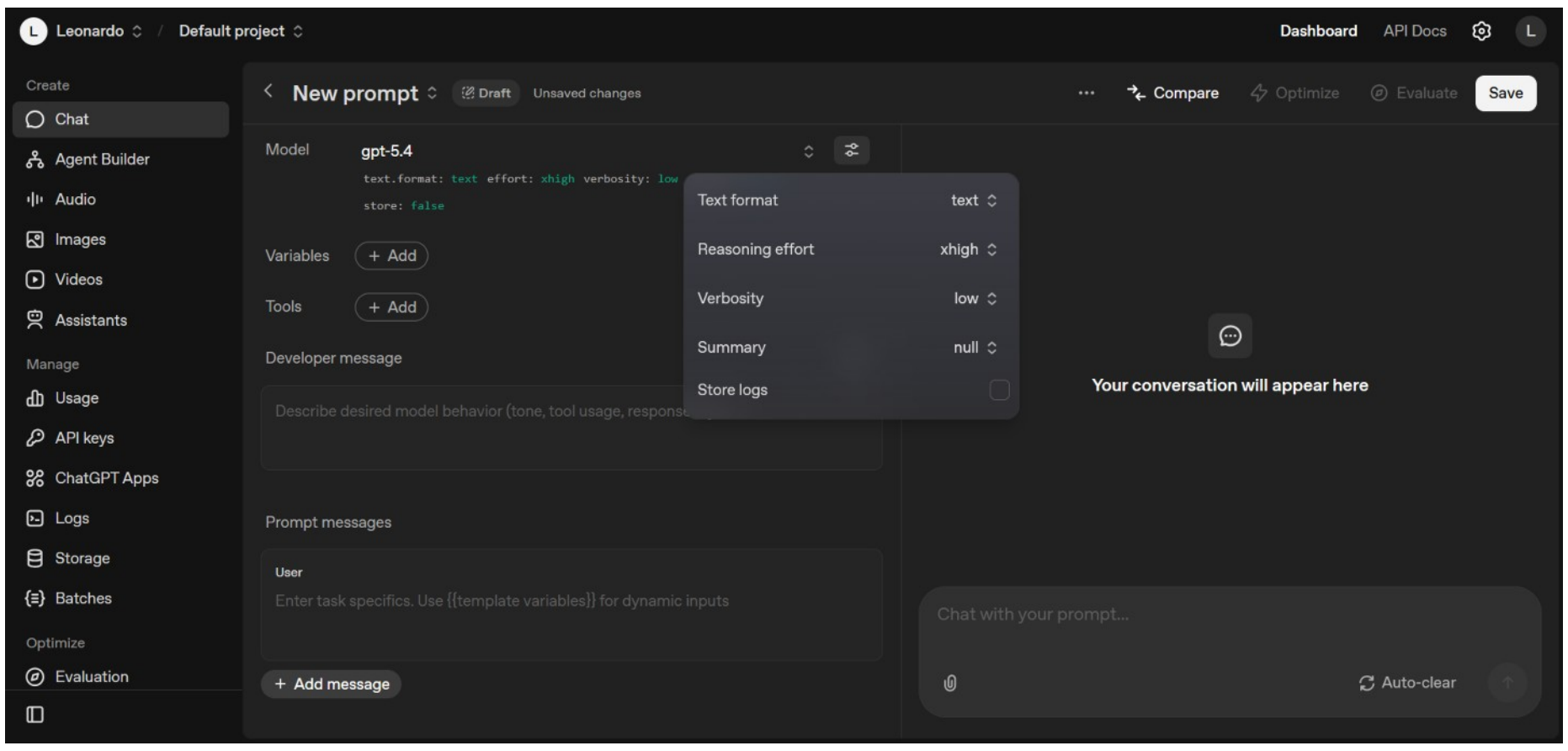

## Prompts for Inductive Content Analysis

### *1. a) Code Generation*

You are an experienced scientist, perfectly educated to conduct text categorization as a qualitative research method. Your task is to analyze the given survey responses and to extract appropriate categories. For that go through the provided survey responses below and identify recurring themes, topics, or intents within the survey responses. Finally, output a table that lists each resulting category alongside a short description (1-2 sentences).

Here are the survey responses:

1

Chat 1

...

n

Chat n

Output the table in Markdown format with one row per category. The first column should contain the found category, and the subsequent column should contain the category description. Example:

| Category | Category Description |
|-------------|------------------------|
| [category] | [category description] |
| [category] | [category description] |
| [category] | [category description] |

…

### ***1. b) Coding***

You are an advanced language model trained for text classification and categorization. Your task is to analyze the given survey responses and assign them to given categories. Assign each message to one or multiple categories, and explain in 1-2 sentences why you decided to do so.

Here are the survey responses:

1

Chat 1

2

Chat 2

…

n

Chat n

Here are the categories:

[list of all categories from 1. a)]

Output a table in Markdown format with one row per message. The first column should contain the chat number (e.g., Chat 1, Chat 2, ..., Chat n), the second column the corresponding category or categories separated by a semicolon, and the subsequent column should contain the categorization explanation. Example:

| Chat Number | Category | Categorization Explanation |

|-------------|-------------|-----------------------|

| Chat 1 | [selected category] | [categorization explanation] |

| Chat 2 | [selected category]; [selected category]; ... | [categorization explanation] |

| Chat 3 | [selected category]; [selected category] | [categorization explanation] |

…

***2. a) Category Generation***

You are an experienced scientist, perfectly educated to conduct text categorization as a qualitative research method. Your task is to analyze given primary categories and to extract appropriate overarching categories. For that go through the provided primary categories below and identify recurring themes, topics, or intents within the primary categories. Finally, output a table that lists each resulting overarching categories alongside a short description (1-2 sentences).

Here are the primary categories:

[list of all categories from 1. a)]

Output the table in Markdown format with one row per overarching category. The first column should contain the found overarching category, and the subsequent column should contain the category description. Example:

| Overarching Category | Category Description |
|-------------|------------------------|
| [overarching category] | [category description] |
| [overarching category] | [category description] |
| [overarching category] | [category description] |

…

***2. b) Grouping Categories***

You are an advanced language model trained for text classification and categorization. Your task is to analyze the given primary categories and assign them to given overarching categories. Assign each primary category to the most relevant overarching category, and explain in 1-2 sentences why you decided to do so.

Here are the primary categories:

[list of all categories from 1. a)]

Here are the overarching categories:

[list of all categories from 2. a)]

Output a table in Markdown format with one row per message. The first column should contain the primary categories (e.g., Primary Category 1, Primary Category 2, ..., Primary

Category n), the second column the corresponding overarching category, and the subsequent column should contain the categorization explanation. Example:

| Primary Category | Overarching Category | Categorization Explanation |

|-------------|-------------|------------------------|

| Primary Category 1 | [selected overarching category] | [categorization explanation] |

| Primary Category 2 | [selected overarching category] | [categorization explanation] |

| Primary Category 3 | [selected overarching category] | [categorization explanation] |

...

## Appendix C - Full Survey Questions

Below for each of the six analysed variables the original full survey question is displayed.

**Table C.1**

| Variable | Survey Question |
|---|---|
| (1) Accessibility Increase | How would you increase accessibility of doctoral education in your country? |
| (2) Financial Situation | How would you improve the financial situation of doctoral students in your country? |
| (3) Well Being | What’s one thing that would most improve doctoral well being in your country? |
| (4) Mobility Barriers | What barriers if any do you face in mobility or recognition of your work? |
| (5) Personal Experience | Is there anything more as a personal experience related to your studies you'd like to share? |
| (6) Anything Else | Is there anything you would like to add about doctoral education in your country from a more general or systemic perspective beyond your personal experience? If relevant you may also mention or cite literature studies or other sources |

related to doctoral education in your country.

---

## Appendix D - Human Coder Opinion on LLM Coding

**(1) Accessibility Increase**

Looks great to me. It indeed grouped ideas together which I think is great and creates an umbrella concept.

**(2) Financial Situation**

The LLM analysis seems fine.

As a general remark, the language model seems to code many variables. I was reluctant to put anything else than "unclear", i.e. it has the tendency to create a (false) signal from full noise, see e.x. variables A1006 and A1239. However, even these fall into the right category even though a proposal is non-existent. My code "possibility to have another job" is often a code it fails to replicate, so one needs to be vigilant when interpreting its code "flexibility to combine work and study".

The financial questions were difficult to evaluate, differentiating between more available grants, larger scholarships and larger salaries seem to point in the same direction even though there are nuanced differences. Similarly, more (system-level) funding and more grants are entirely different things, and I think it [LLM] does an adequate job with the data.

**(3) Well Being**

Overall, I think the LLM coding is reflective of my categorization. I think my primary concerns are the LLM:

1. overinterpreting some of the responses, particularly where they are relatively short/simple and may not be clear (example: ID 192 lacks clarity, yet LLM coded without necessary context)

2. reducing the precision of the coding (example: IDs 313 & 334 treated identically by

the LLM, but not by the human)

3. broadening themes, reducing their usefulness (example: ID 256 reduces multiple human codes and themes into one LLM code and theme that do not fully capture the response)

While I personally note these discrepancies, broadly, the LLM performed well in the subset I've reconsidered, and I would be comfortable with it being used in this manner for the remaining analysis. If my concerns could be noted generally alongside what others find, that would suffice.

**(4) Mobility Barriers**

The idea is the same in both versions. For some questions mine is shorter and concrete however the main idea is the same

**(5) Personal Experience & (6) Anything Else**

In general, the categorization of the AI is in line with my output